\documentclass[10pt,twocolumn,letterpaper]{article}

\usepackage[pagenumbers]{cvpr} %

\usepackage{graphicx}
\usepackage{amsmath}
\usepackage{amssymb}
\usepackage{booktabs}

\usepackage{times}
\usepackage{epsfig}
\usepackage{subcaption}
\usepackage[export]{adjustbox}
\usepackage{makecell}
\usepackage{enumitem}
\usepackage{xcolor}
\usepackage{bm}

\usepackage[pagebackref,breaklinks,colorlinks]{hyperref}

\usepackage[capitalize]{cleveref}
\crefname{section}{Sec.}{Secs.}
\Crefname{section}{Section}{Sections}
\Crefname{table}{Table}{Tables}
\crefname{table}{Tab.}{Tabs.}

\def\cvprPaperID{4855} %
\def\confName{CVPR}
\def\confYear{2022}

\begin{document}

\title{Beyond a Pre-Trained Object Detector: \\Cross-Modal Textual and Visual Context for Image Captioning}

\author{
Chia-Wen Kuo\\
Georgia Tech\\
{\tt\small albert.cwkuo@gatech.edu}
\and
Zsolt Kira\\
Georgia Tech\\
{\tt\small zkira@gatech.edu}
}
\maketitle

\newcommand{\cw}[1]{{\color{blue}{(\textbf{Chia-Wen: }#1)}}}
\newcommand{\zk}[1]{{\color{brown}{(\textbf{Zsolt: }#1)}}}
\newcommand{\js}[1]{{\color{red}{(\textbf{James: }#1)}}}
\newcommand{\nat}[1]{{\color{cyan}{(\textbf{Nathan: }#1)}}}
\newcommand{\jj}[1]{{\color{violet}{(\textbf{Junjiao: }#1)}}}
\newcommand{\amz}[1]{{\color{blue}{(\textbf{Amz: }#1)}}}
\newcommand{\note}[1]{{\color{red}{#1}}}

\renewcommand\cw[1]{}
\renewcommand\zk[1]{}
\renewcommand\js[1]{}
\renewcommand\nat[1]{}
\renewcommand\jj[1]{}
\renewcommand\note[1]{{#1}}

\begin{abstract}
Significant progress has been made on visual captioning, largely relying on pre-trained features and later fixed object detectors that serve as rich inputs to auto-regressive models. A key limitation of such methods, however, is that the output of the model is conditioned only on the object detector's outputs. The assumption that such outputs can represent all necessary information is unrealistic, especially when the detector is transferred across datasets. In this work, we reason about the graphical model induced by this assumption, and propose to add an auxiliary input to represent missing information such as \note{object relationships}. We specifically propose to mine attributes and relationships from the Visual Genome dataset and condition the captioning model on them. Crucially, we propose (and show to be important) the use of a multi-modal pre-trained model (CLIP) to retrieve such contextual descriptions. Further, object detector models are frozen and do not have sufficient richness to allow the captioning model to properly ground them. As a result, we propose to condition both the detector and description outputs on the image, and show qualitatively and quantitatively that this can improve grounding. We validate our method on image captioning, perform thorough analyses of each component and importance of the pre-trained multi-modal model, and demonstrate significant improvements over the current state of the art, specifically +7.5\% in CIDEr and +1.3\% in BLEU-4 metrics.
\end{abstract}
\section{Introduction}\label{section:introduction}

\zk{Overall the paper is well-written! However I think it could use a bit more of an interesting research question emphasis; for example, the CLIP/multi-modal question is not mentioned at all until the end. Start with interesting questions, then describe the answers to them. This should be done consistently throughout. }

For vision-and-language (VL) tasks such as generating textual descriptions of an image (image captioning)~\cite{karpathy2015deep,anderson2018bottom,cornia2020meshed}, it is crucial to encode the input image into a representation that contains relevant information for the downstream language task.
Earlier works use an ImageNet pre-trained model to encode the input image~\cite{vinyals2015show,karpathy2015deep}, while recent works achieve much better performance by using objects detected by an object detector (e.g. Faster R-CNN~\cite{ren2015faster} pre-trained on Visual Genome~\cite{krishnavisualgenome})~\cite{anderson2018bottom,ke2019reflective,qin2019look}.
The detected objects encode more fine-grained information from the input image such as object classes,
locations, and attributes, 
hence achieving substantially better performance.

\begin{figure}
\centering
\includegraphics[width=0.9\linewidth]{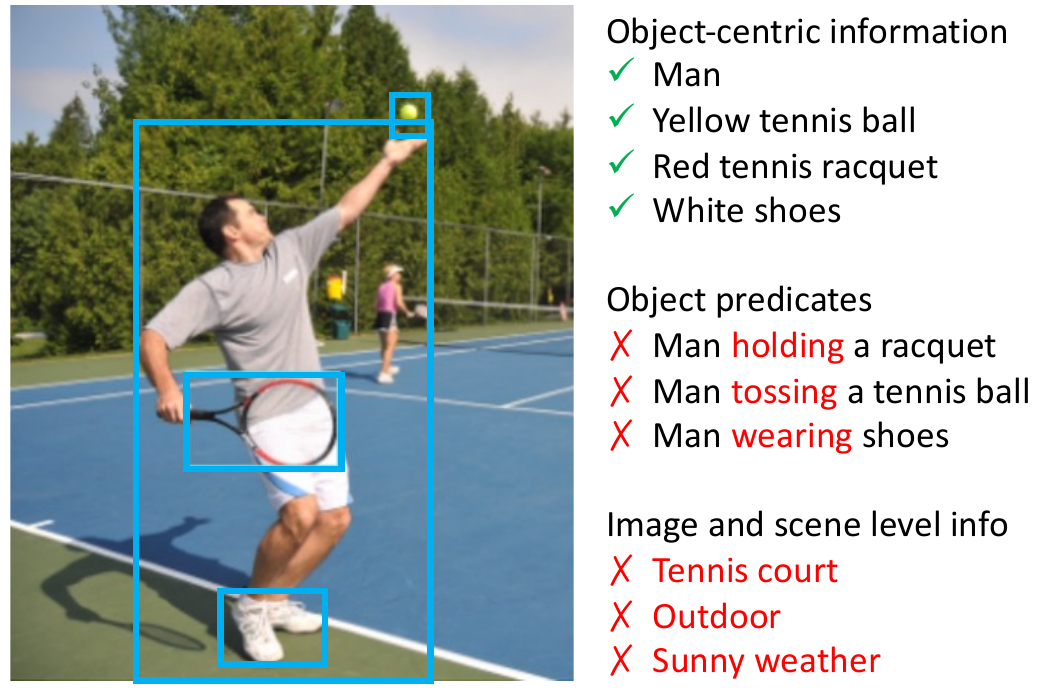}
\caption{
Most existing VL methods encode the input image by a set of objects detected by a frozen pre-trained object detector.
This set of detected objects may be able to provide object-centric information such as object classes, locations, and attributes, but may fail to encode other information also crucial for the target VL tasks such as \note{object predicates}\zk{Since you are training on Visual Genome, wouldn't this be an argument for just using something like scene graphs?} and image/scene level information.
}
\label{figure:concept}
\end{figure}

Despite the success of encoding the input image with detected objects, the object detector is \textit{pre-trained} on datasets such as Visual Genome and kept \textit{frozen} during the training of the target VL task (on a different dataset).
This leads to two major issues as illustrated in Figure~\ref{figure:concept}:
(1) the detector may be good at encoding object-centric information but not at many other kinds of information necessary for the target VL task such as the \note{relationship between objects and image/scene level information};
and
(2) the conditional relationship between the detected objects and the input image is not jointly optimized for the target VL task so that features computed by the object detector cannot
be refined %
before sending into the VL model, potentially resulting in poor features that are difficult to ground, for example.

For (1), most existing works follow prior works~\cite{anderson2018bottom} to pre-train the object detector on Visual Genome for object detection and attribute classification.
This implies that the object features may be good at encoding object-centric information such as classes, locations, and attributes, but not at encoding other crucial information.
Take image captioning as an example; as shown in Figure~\ref{figure:concept}, such crucial information includes \note{relationships between objects (object predicates), image/scene level information, etc.}
Therefore, the first objective of this paper is to provide information complementary to the detected objects.

Inspired by the way the Visual Genome dataset is constructed, we propose to provide complementary but necessary information in the form of contextual text descriptions for image sub-regions.
However, generating the descriptions for image sub-regions requires training another image captioning model, which by itself may not be an easy task.
Therefore, we propose to turn the text generation problem into a \textit{cross-modal retrieval problem}: given an image sub-region, retrieve the top-$k$ most relevant text descriptions from a description database.
One way of doing cross-modal retrieval is to search for visually similar images and return the paired text of that image~\cite{gong2014improving,hodosh2013framing,ordonez2011im2text,sun2015automatic}.
However, we posit that we can effectively leverage 
recent advances in cross-modal pre-training on large-scale image and text pairs, CLIP~\cite{Radford2021LearningTV}, to \textit{directly} retrieve relevant text given an image. %
CLIP has two branches, CLIP-I and CLIP-T which encode image and text, respectively, into a global feature representation, and is trained to pull paired image and text together and push unpaired ones apart.
We show in Section~\ref{section:analysis} that the text descriptions retrieved by CLIP are more relevant to the image query compared to those retrieved indirectly by visual similarity.
The retrieved text descriptions by CLIP provide rich and complementary information,  thus leading to substantial performance improvement.

For (2), in most existing works the pre-trained object detector is kept frozen when training the target VL task.
This implies that the conditioning relationship between the detected objects and the input image is not jointly optimized with the target VL task.
Consequently, the information from a transferred object detector may not result in high-quality features that can be effectively used by the captioning model, for example in grounding them to words. %
Therefore, the second objective of this paper is to strengthen the conditioning relationship between the detected objects and the input image by optimizing this relationship jointly with the target VL task.

To strengthen the conditional relationship for (2), we should first encode the input image into a global feature representation in a way that preserves as much information relevant to the target VL task as possible.
In this paper, we choose CLIP-I, the image branch of CLIP model, as the image encoder.
Since CLIP is also pre-trained on a cross-modal VL task, we show in Section~\ref{section:analysis} that it can better encode information relevant to the target VL tasks compared to models pre-trained on image-only datasets.
We then use a fully connected (FC) layer, which is jointly optimized with the target VL task, to model the conditional relationship.

In this paper, we validate our proposed method on the VL task of image captioning.
By addressing the two issues from using a frozen pre-trained object detector described above, our method improves one of the SoTA image captioning models $\mathcal{M}^2$ by $+7.2\%$ in CIDEr and $+1.3\%$ in BLEU-4.
In summary, we make the following contributions:%
\begin{itemize}[topsep=0pt,itemsep=-1ex,partopsep=1ex,parsep=1ex,labelindent=0.0em,labelsep=0.2cm,leftmargin=*]
\item Identify the potential issues of using detected objects from a frozen pre-trained object detector to encode the input image for image captioning.
\item Propose a cross-modal retrieval module by leveraging the cross-modal joint embedding space by CLIP to retrieve a set of contextual text descriptions that provide information complementary to detected objects.
\item Propose an image conditioning module to strengthen and jointly optimize the conditional relationship between the detected objects and the input image such that the features are more effective and support tasks such as grounding.  %
\item Improve SoTA object-only baseline model by a substantial margin and provide thorough quantitative and qualitative analyses for the proposed two modules and the design choices made within each module.
\end{itemize}

\section{Related Works}

In the early stage of image captioning, researchers used an image encoder such as ResNet~\cite{he2016deep} to encode the input image into a global pooled representation~\cite{vinyals2015show,karpathy2015deep,mao2014deep,donahue2015long,chen2015mind,fang2015captions,jia2015guiding,you2016image,wu2016value,gu2017empirical,chen2017structcap,chen2018groupcap}.
Captions are then generated conditioned on the encoded global feature.
The major issue of using a global pooled representation is that information from the input image is heavily compressed and lost during the encoding process.
For example, all salient objects are fused and spatial information is discarded.
Therefore, follow-on works discard the pooling layer and use grid features from the ConvNet to encode more fine-grained details of the input image~\cite{rennie2017self,xu2015show,lu2017knowing,chen2017sca,jiang2018recurrent}.

In order to further encode more fine-grained details of the input image, Anderson et al.~\cite{anderson2018bottom} propose to encode the input image with a set of objects detected by a frozen object detector.
The object detector is pre-trained on Visual Genome for object detection and attribute classification.
The detected objects are represented by the RoI-pooled features from the object detector.
With the set of detected objects, fine-grained and richer information from the input image such as salient objects, object classes, locations, attributes, etc, are encoded for the downstream VL task, leading to substantial performance improvements.
Because of its great success, encoding the input image with a set of detected objects has become a standard approach in recent VL works~\cite{anderson2018bottom,ke2019reflective,qin2019look,huang2019adaptively,wang2020show,huang2019attention,herdade2019image}, as well as in VL pre-training~\cite{chen2020uniter,tan2019lxmert,lu2019vilbert,su2019vl,li2019visualbert}.

Despite the wide adoption
of encoding the input image with a set of detected objects, as we argued in Section~\ref{section:introduction}, the pre-trained object detector %
may not properly encode other crucial information such as object predicates and image/scene level information as illustrated in Figure~\ref{figure:concept}.
In order to encode more complete information of the input image by a pre-trained object detector, Zhang et al.~\cite{zhang2021vinvl} propose VinVL where the object detector is pre-trained on much larger training corpora that combine multiple public annotated object detection datasets.
Therefore, richer visual objects, attributes, and concepts in the input image are encoded.
Nevertheless, the object detector is still pre-trained to encode object-centric information, %
and other information, e.g. interaction between objects, is not optimized for in the pre-training of the object detector.

To encode information in addition to the detected objects, Li et al.~\cite{li2020oscar} propose OSCAR that includes object tags detected in the image in the form of text for VL pre-training.
Different from our work, the motivation of including object tags in OSCAR is to use them as anchor points to bridge vision and language modalities for VL pre-training.
Also, object predicates are not provided by the object tags.

The image captioning methods discussed above typically generate general descriptions of an input image.
Furthermore, we cannot control the captioning model to generate diverse\zk{Aren't there lots of works targeting improvement of diversity?} or more specific captions that focus on particular objects in the input image.
Therefore, another line of image/video captioning works focus on generating customized captions with additional control signals~\cite{whitehead2018incorporating,chunseong2017attend,park2018towards,rimle2021enriching,attend2u:2017:CVPR}.
This line of works also includes information external of the input image/video such as hashtags~\cite{chunseong2017attend}, or news documents~\cite{whitehead2018incorporating} for caption generation.
Different from standard image captioning methods and our method that train and evaluate the trained model on MS-COCO~\cite{lin2014microsoft}, this line of works
requires additional annotation or labeling during the construction of such custom datasets. 
Our method does not require additional annotations of the input image\zk{Beyond Visual genome annotations and CLIP you mean?} and focuses on  the standard MS-COCO  benchmark.
\section{Method}

\begin{figure}
\centering
\begin{subfigure}[b]{.35\columnwidth}
  \centering
  \includegraphics[width=.8\linewidth]{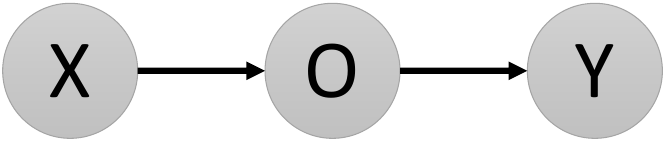}
  \caption{}
  \label{figure:graphical-model-1}
\end{subfigure}%
\begin{subfigure}[b]{.35\columnwidth}
  \centering
  \includegraphics[width=.8\linewidth]{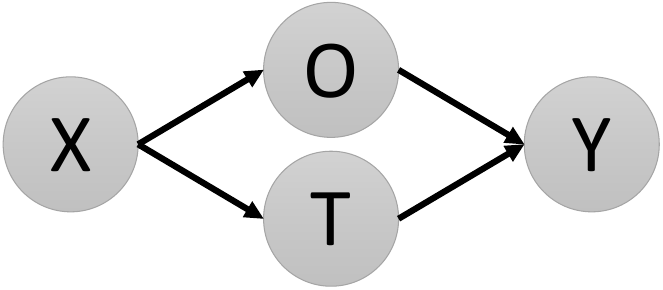}
  \caption{}
  \label{figure:graphical-model-2}
\end{subfigure}
\caption{Graphical models of \textbf{(a)} most existing image captioning models, where $X$ is the input image, $O$ is a set of objects detected by a frozen pre-trained object detector; and \textbf{(b)} our proposed model with a newly introduced node $T$, which represents a set of text descriptions of image sub-regions.}
\label{figure:graphical-model}
\vspace{-3.6mm}
\end{figure}

\subsection{Graphical Model}\label{section:graphical-model}

\begin{figure*}
\centering
\includegraphics[width=0.9\linewidth]{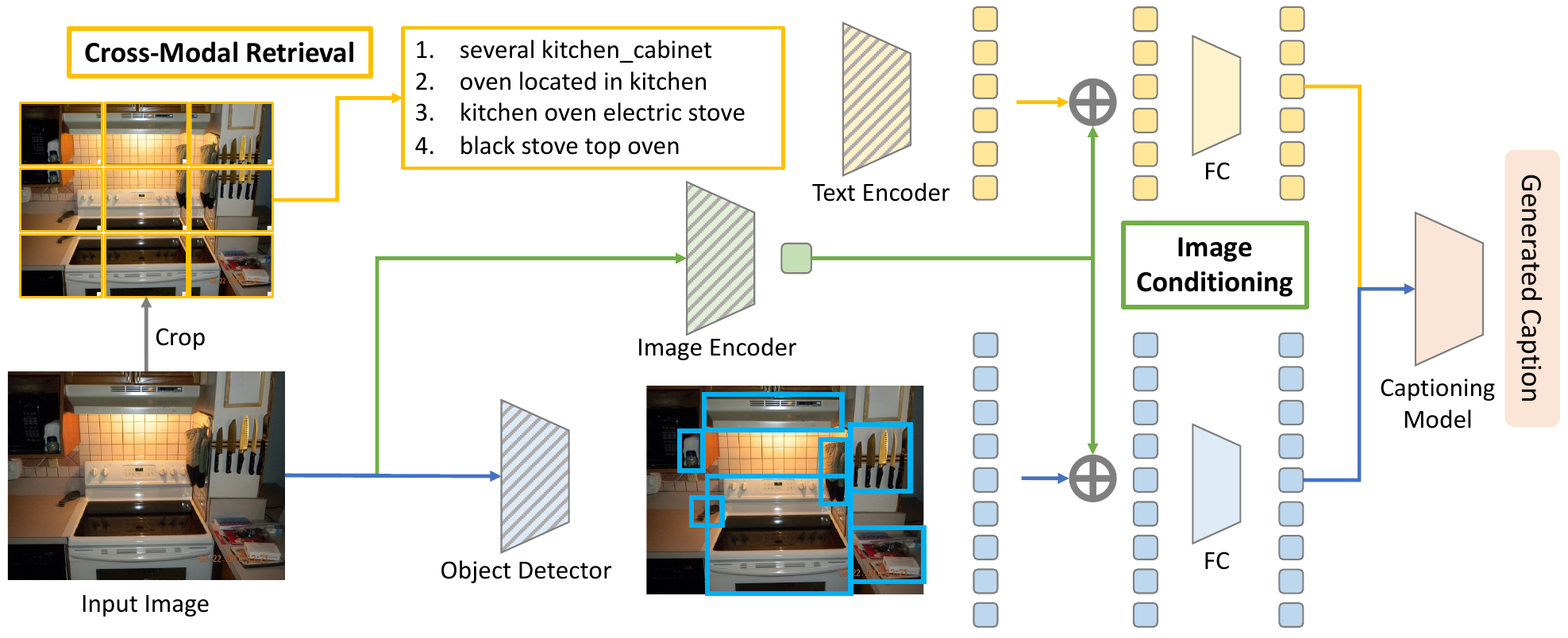}
\caption{
Model architecture.
We propose (1) a cross-modal retrieval module to retrieve a set of contextual text descriptions that provide information complementary to the detected objects as shown in the yellow box.
We also propose (2) an image conditioning module to strengthen the conditional relationship between the detected objects and the input image as shown in the green box.
The models with shaded patterns (text encoder, image encoder, and object detector) are pre-trained and kept frozen.
Only the FCs and the captioning model are trained for the target VL task.
The $\bigoplus$ symbol represents concatenation operation along the feature dimension.
Each token ($\square$ symbol) represents a $d$-dimensional feature vector.
The image features (green token) are broadcast before the concatenation operation.
\js{The font for ``generated caption" is *very* difficult to read}
}
\label{figure:model}
\end{figure*}

Most existing works model the image captioning problem with a graphical model shown in Figure~\ref{figure:graphical-model-1}, where given an input image $X$, a set of objects $O$ are detected by a frozen pre-trained object detector, and the caption $Y$ is generated conditioned on $O$.
The graphical model with the chain structure shown in Figure~\ref{figure:graphical-model-1} can be derived as:
\begin{align}
p(\bm{y}|\bm{x}) &= \prod_i p(y_i|\bm{x},y_{1:i-1}) \nonumber \\
&= \prod_i \sum_O p(\bm{o}|\bm{x},y_{1:i-1}) p(y_i|\bm{x},\bm{o},y_{1:i-1}) \nonumber \\
&= \prod_i \sum_O p(\bm{o}|\bm{x}) p(y_i|\bm{x},\bm{o},y_{1:i-1}) \label{equation:image-condition-1} \\
&= \prod_i \sum_O p(\bm{o}|\bm{x}) p(y_i|\bm{o},y_{1:i-1}) \label{equation:image-condition-2} \\
&\simeq \prod_i p(y_i|\bm{o},y_{1:i-1}) \label{equation:image-condition-3}
\end{align}
where $p(y_i|\bm{o},y_{1:i-1})$ is modeled as an auto-regressive caption generation model.
Between Equation~\ref{equation:image-condition-1} and Equation~\ref{equation:image-condition-2}, it is assumed that $\bm{o}$ completely encode all necessary information of $\bm{x}$ so that $y_i$ is conditionally independent of $\bm{x}$.
Between Equation~\ref{equation:image-condition-2} and Equation~\ref{equation:image-condition-3}, researchers typically take the argmax and threshold to select a fixed set of detected objects from the object detector.
From the graphical model derived above, we can clearly identify the two major issues arising from a frozen pre-trained object detector.
First, the assumption that $\bm{o}$ completely encodes all necessary information of $\bm{x}$.
In practice, the object detector pre-trained on Visual Genome for object detection and attribute prediction may fail to encode crucial information of $\bm{x}$ such as the relationship between objects and image/scene level information.
Second, the conditional relationship between the detected objects $\bm{o}$ and the input image $\bm{x}$ is computed by a frozen pre-trained object detector and is not jointly optimized with the target image captioning task.
Therefore, the features computed by the frozen pre-trained object detector cannot be refined before sending into the auto-regressive caption generation model, leading to potentially poor features especially given that they are trained on a different dataset. %

To mitigate issue (1), one na\"ive solution is to pre-train the object detector to predict other information such as the predicates between objects so that more complete information can be encoded by $\bm{o}$.
However, it is still an open research question to effectively train a network to model the interaction between objects, especially across datasets~\cite{yang2018graph,xu2017scene,li2017scene,zellers2018neural,tian2020image}.
Therefore, in this paper, we propose to insert another node $T$ into the model, as shown in Figure~\ref{figure:graphical-model-2}, to encode information complementary to $O$ without re-training the object detector.
By including both $O$ and $T$, more complete information of $X$ is encoded, and thus the conditional independence assumption between Equation~\ref{equation:image-condition-1} and Equation~\ref{equation:image-condition-2} are better supported.
We can similarly derive the graphical model with the newly introduced node $T$:
\begin{equation}
p(\bm{y}|\bm{x}) \simeq \prod_i p(y_i|\bm{o}, \bm{t},y_{1:i-1}) \label{equation:new-model}
\end{equation}
Please see supplementary for the full derivation.

To mitigate issue (2), we propose to use a fully connected (FC) layer to refine the features of each detected object conditioned on the features of input image $X$.
The FC layer is jointly optimized with the training objective of the image captioning task in order to strengthen the conditional relationship between $O$ and $X$, and we show such feature refinement can lead to features supporting qualitatively and quantitatively improved grounding and  results. %

We illustrate the overall model in Figure~\ref{figure:model}.
To address issue (1), we introduce a cross-modal retrieval module (yellow box) to retrieve a set of text descriptions $T$ that encode information complementary to detected objects $O$ from the input image.
To address issue (2), we introduce an image conditioning module (green box) to strengthen the conditional relationship between the detected object and the input image.
Key to our method is that both of these approaches will allow us to leverage large-scale cross-modal models that have recently been introduced. In the rest of this section, we describe how to obtain $T$ in Section~\ref{section:text} and how to model the conditional relationship in Section~\ref{section:image}.

\subsection{Text Descriptions} \label{section:text}

\begin{figure*}[t]
\begin{subfigure}[b]{.32\linewidth}
  \centering
  \includegraphics[width=.9\linewidth]{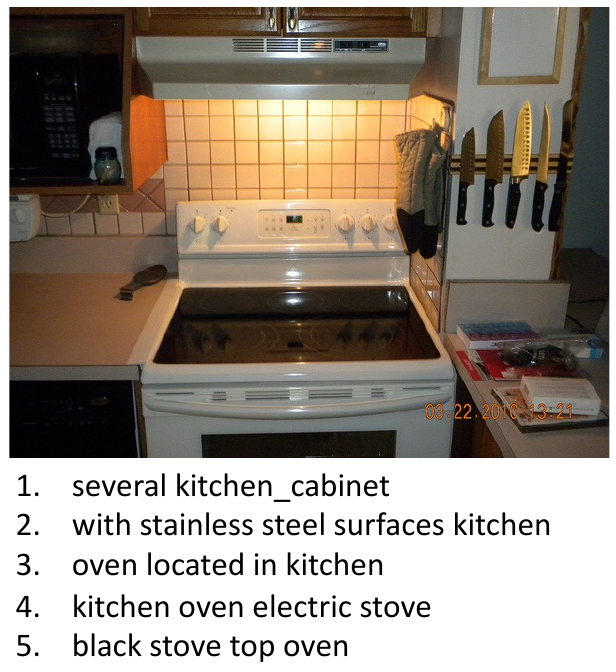}
  \caption{Original image and the retrieved top-5\\ most relevant text descriptions.}
  \label{figure:original}
\end{subfigure}
\begin{subfigure}[b]{.32\linewidth}
  \centering
  \includegraphics[width=.9\linewidth]{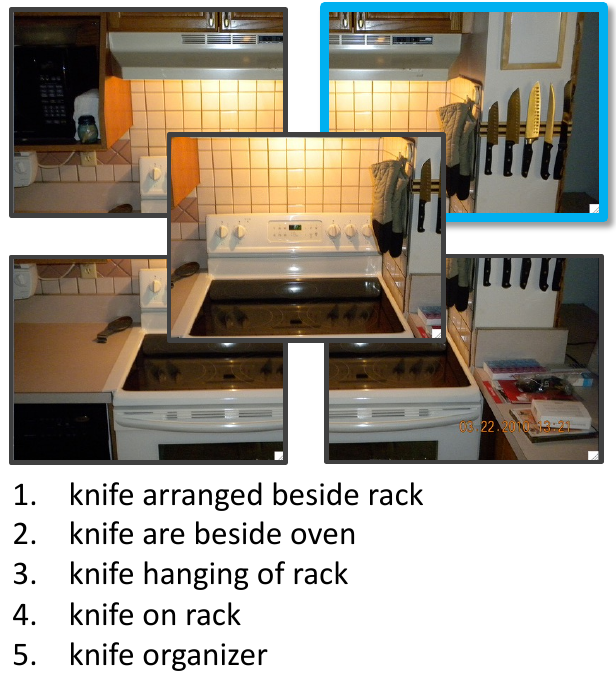}
  \caption{Image five crops and the retrieved top-5\\ text descriptions for the crop in the blue box.}
  \label{figure:five}
\end{subfigure}
\begin{subfigure}[b]{.32\linewidth}
  \centering
  \includegraphics[width=.9\linewidth]{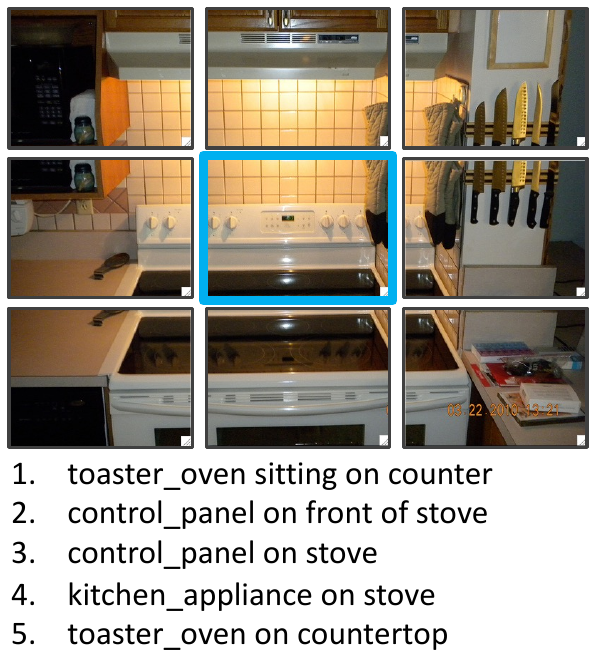}
  \caption{Image nine crops and the retrieved top-5\\ text descriptions for the crop in the blue box.}
  \label{figure:nine}
\end{subfigure}
\caption{Retrieved top-5 most relevant text descriptions for \textbf{(a)} the original image, \textbf{(b)} the image five crops, and \textbf{(c)} the image nine crops.
For five crops and nine crops, we show the retrieved text descriptions for the crop in the blue box.\js{If time, see if you can make word wrap on subcaptions move to next line sooner, so that the sub captions are not so close together when not necessary}}
\label{figure:image-subregions}
\end{figure*}

In the last section, we introduce $T$, a set of text descriptions providing information complementary to the detected objects $O$.
Imagine when one is asked to describe an image, he/she may first focus on local regions of the image and then gradually merge the local information to generate the final description of the whole image.
Similarly, we propose to generate text descriptions for image sub-regions as shown in Figure~\ref{figure:image-subregions} so that those descriptions contain more details and provide more complete information of the input image that can be merged in the later stage.
Instead of training another captioning model for generating descriptions of image sub-regions, which by itself may not be an easy task, we propose to retrieve top-$k$ most relevant descriptions from a description database for each image sub-region, thus turning this into a cross-modal retrieval problem.
We describe the three steps for cross-modal retrieval as follows.

\textbf{Description database}
The descriptions database is the source of relevant text descriptions for an image sub-region, and we choose the top-$k$ most relevant ones.
In this paper, we propose to parse the annotations from the Visual Genome dataset (which is already commonly used to train object detectors on) to construct the description database.
Instead of taking the \textit{region descriptions} from Visual Genome, which contain many similar sentences (and we show to be inferior in experiments), we parse the annotations of \textit{attributes} and \textit{relationships}.
Specifically, the attribute annotations take the form of ``attribute-object'' pairs.
We first convert the object name to its synset canonical form and then collect all the ``attribute-object'' pairs.
On the other hand, the relationship annotations take the form of ``subject-predicate-object'' triplets.
We similarly convert the subject and object names to their synset canonical forms and then collect all the ``subject-predicate-object'' triplets.
Finally, we merge all the collected ``attribute-object'' pairs and ``subject-predicate-object'' triplets and remove duplicates to construct the descriptions database.

\textbf{Text description retrieval}. Our goal is to retrieve the top-$k$ most relevant text descriptions from the description database given a query of an image sub-region.
This involves two sub-problems: (1) how to generate image sub-regions and (2) how to perform cross-modal retrieval between image and text.
For (1), we propose to generate the five crops (Figure~\ref{figure:five}) and nine crops (Figure~\ref{figure:nine}) of the original image.
These crops may contain multiple objects rather than just a single salient object, which is beneficial for capturing the interaction between objects if we are able to retrieve good text descriptions for the crop.
For (2), we propose to leverage the cross-modal joint embedding from CLIP~\cite{Radford2021LearningTV} for this cross-modal retrieval problem.
The CLIP model has two branches: the image branch CLIP-I and text branch CLIP-T that encode image and text into a global feature representation,  respectively.
CLIP is trained on large-scale image and text pairs such that paired image and text are pulled together in the embedding space while unpaired ones are pushed apart.
With the pre-trained CLIP model, the cross-modal retrieval problem becomes a nearest neighbor search in CLIP's cross-modal embedding space.
Specifically, we use CLIP-T to encode all text descriptions in the description database as the searching keys.
The image sub-region from five crops and nine crops as well as the original image is encoded by CLIP-I into a query.
We then search in the description database for the text descriptions with the top-$k$ highest cosine similarity scores.
Therefore, we will have a set of retrieved text descriptions $T=\{t_{i,j,k}|i\in\{\text{original, five, nine}\}, j\in\{1,2,...,\text{\#crops}_i\}, k\in\{1,2,...,\text{top-}k\}\}$, where subscript $i$ represents whether it is from the original image, five crop, or nine crop sets; subscript $j$ represents the $j$-th crop for $t_i$ (e.g. top-left, bottom right, etc. for five crop); $\text{\#crops}_\text{\{original, five, nine\}}$ equals to \{1, 5, 9\}, respectively; and subscript $k$ represents the top-$k$-th retrieval.
Some examples of the top-5 results are shown in Figure~\ref{figure:image-subregions}.

\textbf{Text encoding}.
After retrieving the set of text descriptions $T=\{t_{i,j,k}\}$, we use a pre-trained text encoder to encode each of them into a global representation.
In this paper, we use a frozen pre-trained CLIP-T as the text encoder as CLIP is similarly pre-trained on a VL task so that it could better encode relevant information for the target VL task from the retrieved text descriptions.
The three steps described above, from constructing the description database to searching it and finally encoding the retrieved text descriptions for $T$, can be done offline for each image in the benchmark dataset in the same way as the detected objects $O$.
To further distinguish between different $i$ (original, five crop, or nine crop) and $j$ ($j$-th crop for $t_i$), we add a learnable embedding to $t_{i,j,k}$ for different $i$ and $j$.

\subsection{Image Conditioning} \label{section:image}

In Section~\ref{section:graphical-model}, we propose to model and strengthen the conditional relationship between the detected object $O$ and the input image $X$ so that the features computed by the object detector can be refined before sending into the captioning model.
Since the text descriptions are also retrieved offline by the pre-trained CLIP model, we similarly want to strengthen the conditional relationship between the retrieved text descriptions $T$ and the input image $X$.
As shown in the green box of Figure~\ref{figure:model}, we propose to condition each detected object and retrieved text description on the input image, and model this conditional relationship by fully connected (FC) layers.

To condition the detected objects $\bm{o}$ and the retrieved text descriptions $\bm{t}$ on the input image $\bm{x}$, we first encode $\bm{x}$ into a global representation $f_x \in \mathbb{R}^{d_x}$.
We require that the encoded representation for $\bm{x}$ preserves as much information relevant to the target VL task as possible.
In this paper, we use a frozen pre-trained CLIP-I as the image encoder as it is similarly pre-trained on a VL task such that it could better encode information relevant to the target VL task from the input image.
We use the following notations: $\bm{o} = \{o_1, o_2,...,o_n\}$, where each $o \in \mathbb{R}^{d_o}$, is a set of objects detected by a frozen pre-trained object detector;
and $\bm{t} = \{t_{i,j,k}|\forall i,j,k\}$, where each $t_{i,j,k} \in \mathbb{R}^{d_t}$, is the set of retrieved text descriptions encoded by CLIP-T (described in Section~\ref{section:text}).
We then model the conditional relationship as:
\begin{equation}
\begin{aligned}
\hat{o}_{m} &= \text{drop}(\text{fc}_o(\text{norm}_o([o_m, f_x]))) \\
\hat{t}_{i,j,k} &= \text{drop}(\text{fc}_{t_i}(\text{norm}_{t_i}([t_{i,j,k}, f_x]))),
\end{aligned}
\end{equation}
where $[\cdot,\cdot]$ is concatenation along the feature dimension, \textit{norm} is a layer normalization layer, and \textit{drop} is a dropout layer.
Note that we encode each $\bm{t}_i$ separately with different FC layers and norm layers as those are text descriptions retrieved for different granularity (original, five crops, or nine crops).
Finally, we collect the image conditioned sequences: $\hat{\bm{o}}=\{\hat{o}_1, \hat{o}_2,...,\hat{o}_n\}$, $\hat{\bm{t}}_i=\{\hat{t}_{i,j,k}|\forall j,k\}$.

\subsection{Image Captioning}

Incorporating the image conditioned objects $\hat{\bm{o}}$ and text descriptions $\hat{\bm{t}}_i$ into an image captioning model is simple.
As shown in Equation~\ref{equation:image-condition-1}, an image captioning model is typically an auto-regressive model $p(y_i|\bm{o},y_{1:i-1})$, which takes a sequence of detected objects $\bm{o}$ as input.
Therefore, without modifying the image captioning model, we simply need to concatenate the image conditioned objects $\hat{\bm{o}}$ and text descriptions $\hat{\bm{t}}_i$ along the sequence dimension as $\bm{z} = [\hat{\bm{o}}, \hat{\bm{t}}_\text{original}, \hat{\bm{t}}_\text{five}, \hat{\bm{t}}_\text{nine}]$, and feed $\bm{z}$ into it in place of $\bm{o}$ as $p(y_i|\bm{z},y_{1:i-1})$.
The model can then be trained with the commonly used maximum log-likelihood loss for word prediction and fine-tuned with the RL loss using CIDEr score as reward~\cite{rennie2017self,cornia2020meshed} in the same way as before.

\section{Experiments}

\subsection{Implementation Details}
In this paper, we incorporate our method into one of the state-of-the-art image captioning models, $\mathcal{M}^2$~\cite{cornia2020meshed}, and train and evaluate our method on the Karpathy split~\cite{karpathy2015deep} of the MS-COCO dataset~\cite{lin2014microsoft}.
We tune the top-$k$ parameter on the validation set and find that the performance saturates at $k=12$.
Therefore, we set $k=12$ across all the experiments.
In our proposed model shown in Figure~\ref{figure:model}, the image encoder CLIP-I and text encoder CLIP-T are both frozen.
Only the FC layers are trainable, which contains one order fewer parameters compared to the image captioning model.

\subsection{Main Results}
\begin{table}
\centering
\renewcommand{\arraystretch}{1.2}
\resizebox{1.0\columnwidth}{!}{
\begin{tabular}{@{\extracolsep{4pt}}lcccccc@{}}
\toprule
Method & B-1 & B-4 & M & R & C & S \\
\midrule
SCST~\cite{rennie2017self} & - & 34.2 & 26.7 & 55.7 & 114.0 & - \\
Up-Down~\cite{anderson2018bottom} & 79.8 & 36.3 & 27.7 & 56.9 & 120.1 & 21.4 \\
RFNet~\cite{jiang2018recurrent} & 79.1 & 36.5 & 27.7 & 57.3 & 121.9 & 21.2 \\
HIP~\cite{yao2019hierarchy} & - & 38.2 & 28.4 & 58.3 & 127.2 & 21.9 \\
GCN-LSTM~\cite{yao2018exploring} & 80.5 & 38.2 & 28.5 & 58.3 & 127.6 & 22.0 \\
SGAE~\cite{yang2019auto} & 80.8 & 38.4 & 28.4 & 58.6 & 127.8 & 22.1 \\
ORT~\cite{herdade2019image} & 80.5 & 38.6 & 28.7 & 58.4 & 128.3 & 22.6 \\
AoANet~\cite{huang2019attention} & 80.2 & 38.9 & 29.2 & 58.8 & 129.8 & 22.4 \\
$\mathcal{M}^2$~\cite{cornia2020meshed} & 80.8 & 39.1 & 29.1 & 58.4 & 131.2 & 22.6 \\
$\mathcal{M}^{2\dagger}$ & 80.2 & 38.4 & 29.1 & 58.4 & 128.7 & 22.9 \\
\midrule
$\mathcal{M}^{2\dagger}$ + Ours & \textbf{81.5} & \textbf{39.7} & \textbf{30.0} & \textbf{59.5} & \textbf{135.9} & \textbf{23.7} \\
\bottomrule
\end{tabular}
}
\caption{
Image captioning results on the test set of MS-COCO Karpathy split\cite{karpathy2015deep}.
We incorporate our proposed method into the baseline image captioning model $\mathcal{M}^2$ using their released code.
For fair comparison, we also show the performance of the released code (denoted as $\mathcal{M}^{2\dagger}$), which is slightly lower than those reported in the paper.
}
\label{table:main}
\end{table}

\begin{table}
\centering
\renewcommand{\arraystretch}{1.2}
\resizebox{\linewidth}{!}{
\begin{tabular}{@{\extracolsep{4pt}}ccccccc@{}}
\toprule
Model & \makecell{Detector\\Pre-training} & \makecell{Transformer\\Pre-training} & B-4 & C & S \\
\midrule
$\mathcal{M}^{2\dagger}$ & VinVL~\cite{zhang2021vinvl} & None & 40.5 & 135.9 & 23.5 \\
$\mathcal{M}^{2\dagger}$ + Ours & VinVL & None & \textbf{41.4} & \textbf{139.9} & \textbf{24.0} \\
\hline
OSCAR~\cite{li2020oscar} & VG~\cite{krishnavisualgenome} & 6.5M & 40.5 & 137.6 & 22.8 \\
VinVL~\cite{zhang2021vinvl} & VinVL & 8.85M & 40.9 & 140.4 & \textbf{25.1} \\
OSCAR + Ours & VG & 6.5M & \textbf{41.3} & \textbf{142.2} & 24.9 \\
\bottomrule
\end{tabular}
}
\caption{
Detector pre-training \textit{v.s.} transformer pre-training.
When combined with other advanced pre-training techniques, our proposed method achieves competitive performance against VinVL~\cite{zhang2021vinvl}, state-of-the-art method with \textit{both} large-scale detector and transformer pre-training.
}
\label{table:SoTA}
\vspace{-1mm}
\end{table}

We first compare with the trained-from-scratch methods in Table~\ref{table:main}.
We show the results on the test set with cross-entropy training and then SCST~\cite{rennie2017self} RL fine-tuning.
With complementary information provided by the retrieved text descriptions and image conditioning, our method improves the baseline model $\mathcal{M}^{2\dagger}$ by $+7.2\%$ in CIDEr and $+1.3\%$ in BLEU-4, and compares favorably with all previous trained-from-scratch methods across all metrics.

We then compare with the methods with more advanced pre-training techniques in Table~\ref{table:SoTA}, and show that when combined together, our method performs competitively against state-of-the-art VinVL~\cite{zhang2021vinvl}.
Specifically, VinVL improves the object features by pre-training a larger object detector model on large training corpora that combine multiple object detection datasets, as opposed to the conventional approaches that pre-train the object detector on the Visual Genome (VG) dataset.
On the top half of Table~\ref{table:SoTA}, we can see that our method is able to provide information complementary to VinVL pre-trained detector and achieves better performance compared with $\mathcal{M}^{2\dagger}$ + VinVL detector.
On the other hand, recent methods (\textit{e.g.} OSCAR~\cite{li2020oscar} and VinVL~\cite{zhang2021vinvl}) propose to pre-train the cross-modal transformer on large corpora of image-caption pairs and achieve SoTA performance.
On the lower half of Table~\ref{table:SoTA}, by combining our method with OSCAR~\cite{li2020oscar}, our method achieves competitive performance compared to VinVL, which requires \textit{both} large-scale detector pre-training and transformer pre-training.
Lastly, by comparing OSCAR and OSCAR+Ours, we verify our claim that our proposed method indeed provides information in addition to objects, as OSCAR explicitly includes object tags of the image as part of the inputs to the model.

\subsection{Analysis}\label{section:analysis}

To verify the effectiveness of the proposed text description module and the image conditioning module, as well as the design choices made within each module, we provide detailed analyses in this section.

\begin{table}
\centering
\renewcommand{\arraystretch}{1.2}
\resizebox{0.8\columnwidth}{!}{
\begin{tabular}{@{\extracolsep{4pt}}cccccccc@{}}
\toprule
Text & Image & B-1 & B-4 & C & S \\
\midrule
& & 75.74 & 35.47 & 112.39 & 20.41 \\
& \checkmark & 77.33 & 36.96 & 116.84 & 21.41 \\
\checkmark & & 77.07 & 37.12 & 116.99 & 21.30 \\
\checkmark & \checkmark & \textbf{77.45} & \textbf{37.74} & \textbf{118.87} & \textbf{21.45} \\
\bottomrule
\end{tabular}
}
\caption{
Ablation study for the proposed text descriptions (denoted as \textit{Text}) and image conditioning (denoted as \textit{Image}).
The first row without \textit{Text} and \textit{Image} corresponds to the baseline model using only detected objects for encoding the input image.
}
\label{table:ablation}
\end{table}

\textbf{Ablation study}. We ablate the two major components proposed in this paper: (1) text descriptions and (2) image conditioning.
We use the $\mathcal{M}^2$ model trained with cross-entropy loss as the baseline model, which uses only the detected objects to encode the input image.
Results are shown in Table~\ref{table:ablation}.
We can see that adding either one of text descriptions or image conditioning brings substantial performance improvement compared to the baseline model.
Overall, with these two modules combined, our proposed method achieves $+6.5\%$ performance improvement in CIDEr and $+2.3\%$ in BLEU-4.
This means the proposed two components are indeed capable of providing information complementary to the detected objects, and the complementary information is beneficial for image captioning.
Qualitatively, we show some top-5 retrieved text descriptions in Figure~\ref{figure:image-subregions}, where we can see that the retrieved text descriptions well describe the image sub-regions and provide complementary information such as object predicates.

\begin{figure}
\centering
\includegraphics[width=1\linewidth]{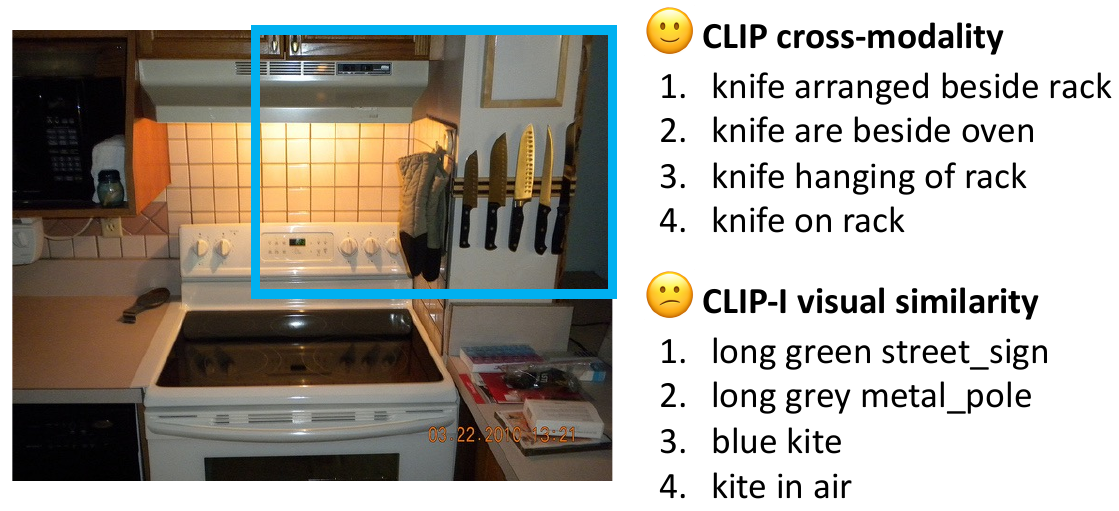}
\caption{
Retrieved text descriptions for the image sub-region in the blue box from \textit{\textbf{(top)}} cross-modal joint embedding by CLIP \textit{v.s.} \textit{\textbf{(bottom)}} visual similarity by CLIP-I.
}
\label{figure:retrieval}
\end{figure}

\begin{table}
\centering
\renewcommand{\arraystretch}{1.2}
\resizebox{0.9\columnwidth}{!}{
\begin{tabular}{@{\extracolsep{4pt}}cccccc@{}}
\toprule
Encoder & B-1 & B-4 & C & S \\
\midrule
R-101~\cite{he2016deep} & 75.82 & 36.27 & 112.77 & 20.54 \\
ViT~\cite{dosovitskiy2020vit} & 76.00 & 35.13 & 112.53 & 20.52 \\
MoCo-v2~\cite{chen2020mocov2} & 76.03 & 35.35 & 112.06 & 20.51 \\
CLIP-I~\cite{Radford2021LearningTV} & 75.87 & 36.52 & 113.66 & 20.66 \\
\midrule
Object only baseline & 75.74 & 35.47 & 112.39 & 20.41 \\
Ours (CLIP) & \textbf{77.07} & \textbf{37.12} & \textbf{116.99} & \textbf{21.30} \\
\bottomrule
\end{tabular}
}
\caption{
Retrieving text descriptions by visual similarity with different image encoders.
}
\label{table:design-text}
\end{table}
\textbf{Text descriptions retrieval by visual similarity}.
In Section~\ref{section:text}, we propose to leverage the cross-modal joint embedding from CLIP to retrieve text descriptions for each image crop.
Another common way of performing cross-modal retrieval is to search for visually similar images and return the paired text of the image~\cite{gong2014improving,hodosh2013framing,ordonez2011im2text,sun2015automatic}.
In the analysis, we first encode the image crop as query and encode the bounding boxes (bboxes) annotations from Visual Genome as key with an image encoder (either R-101~\cite{he2016deep}, MoCo-v2~\cite{chen2020mocov2}, ViT~\cite{dosovitskiy2020vit} or CLIP-I~\cite{Radford2021LearningTV}).
Then the top-$k$ bboxes with the highest cosine similarity scores are selected and the text descriptions associated with the selected bboxes are returned as the retrieved text descriptions.
Finally, we train the $\mathcal{M}^2$ model with detected objects and the retrieved text descriptions.
We can see in Table~\ref{table:design-text} that incorporating text descriptions retrieved by visual similarity does not bring much performance improvement compared to the object-only baseline.
Qualitatively, we can see in Figure~\ref{figure:retrieval} that many of the retrieved text descriptions are irrelevant to the query image.
On the other hand, by leveraging the cross-modal joint embedding by CLIP, the retrieved text descriptions are highly relevant to the query image.
Therefore, the retrieved text descriptions can provide complementary and relevant information, which is beneficial. %

\textbf{How to retrieve text descriptions?}
\begin{table}
\begin{minipage}{.49\linewidth}
\centering
\renewcommand{\arraystretch}{1.2}
\resizebox{0.85\linewidth}{!}{
\begin{tabular}{@{\extracolsep{4pt}}cccc@{}}
\toprule
\makecell{Caption\\dataset} & B-4 & C & S \\
\midrule
- & 35.47 & 112.39 & 20.41 \\
COCO & \textbf{37.37} & \textbf{117.17} & 21.19 \\
CC & 37.30 & 117.13 & 21.14 \\
VG (Ours) & 37.12 & 116.99 & \textbf{21.30} \\
\bottomrule
\end{tabular}
}
\caption{
Text descriptions retrieved from different image captioning datasets.
}
\label{table:text-database}
\end{minipage}
\begin{minipage}{.02\linewidth}
\centering
\renewcommand{\arraystretch}{1.2}
\resizebox{1.0\linewidth}{!}{
\begin{tabular}{@{\extracolsep{4pt}}c@{}}
\toprule
\bottomrule
\end{tabular}
}
\end{minipage}
\begin{minipage}{.46\linewidth}
\centering
\renewcommand{\arraystretch}{1.2}
\resizebox{0.85\linewidth}{!}{
\begin{tabular}{@{\extracolsep{4pt}}cccc@{}}
\toprule
Query & B-4 & C & S \\
\midrule
- & 35.47 & 112.39 & 20.41 \\
Whole  & 36.39 & 115.98 & 20.94 \\
Five & 37.04 & 116.73 & 21.21 \\
Nine & 36.95 & 116.03 & 21.10 \\
All (Ours) & \textbf{37.12} & \textbf{116.99} & \textbf{21.30} \\
\bottomrule
\end{tabular}
}
\caption{
Text descriptions retrieved by using different image crops as query.
}
\label{table:text-retrieval}
\end{minipage}
\end{table}
In Section~\ref{section:text}, we describe how to construct the description database where the text descriptions are retrieved from, and how to retrieve text descriptions by image crops (see Figure~\ref{figure:image-subregions}).
In this subsection, we answer the following questions: (1) Is the final performance sensitive to the description database? (2) How effective is the proposed text retrieval method by using different image crops as query?
We train the $\mathcal{M}^2$ model with detected objects and the text descriptions retrieved with different methods.
For (1), we construct the description databse on different image captioning datasets including Visual Genome~\cite{krishnavisualgenome}, MS-COCO~\cite{lin2014microsoft}, and Conceptual Captions~\cite{sharma2018conceptual}.
In Table~\ref{table:text-database}, we show that the performance is \textit{not} sensitive to the description databases the text descriptions are retrieved from.
For (2), we ablate the text retrieval strategies of using the whole image, five crops, nine crops or the combination of all the above in Table~\ref{table:text-retrieval}.
We can see that using image crops to retrieve more fine-grained text descriptions is beneficial.
Qualitative results of the retrieved fine-grained text descriptions using different image crops are shown in Figure~\ref{figure:image-subregions}.

\textbf{Modeling of image conditioning}.
\begin{table}
\begin{minipage}{.465\linewidth}
\centering
\renewcommand{\arraystretch}{1.2}
\resizebox{1.0\linewidth}{!}{
\begin{tabular}{@{\extracolsep{4pt}}ccccc@{}}
\toprule
\makecell{Conditioning\\Method} & \#Tokens & B-4 & C & S \\
\midrule
- & 50 & 35.47 & 112.39 & 20.41 \\
TF-V & 51 & 36.66 & 116.01 & 21.16 \\
TF-G & 100 & 36.75 & 116.22 & 21.36 \\
FC (Ours) & 50 & \textbf{36.96} & \textbf{116.84} & \textbf{21.41} \\

\bottomrule
\end{tabular}
}
\caption{
Different image conditioning methods. See text for detailed descriptions of TF-V and TF-G. We use CLIP-I to encode the input image.
}
\label{table:img-cond-method}
\end{minipage}
\begin{minipage}{.02\linewidth}
\centering
\renewcommand{\arraystretch}{1.2}
\resizebox{1.0\linewidth}{!}{
\begin{tabular}{@{\extracolsep{4pt}}c@{}}
\toprule
\bottomrule
\end{tabular}
}
\end{minipage}
\begin{minipage}{.497\linewidth}
\centering
\renewcommand{\arraystretch}{1.2}
\resizebox{1.0\linewidth}{!}{
\begin{tabular}{@{\extracolsep{4pt}}ccccc@{}}
\toprule
Encoder & \makecell{Pre-trained\\Dataset} & B-4 & C & S \\
\midrule
R-101~\cite{he2016deep} & IN-1K~\cite{imagenet_cvpr09} & 35.64 & 113.20 & 21.00 \\
BiT~\cite{kolesnikov2020big} & JFT-300M~\cite{sun2017revisiting} & 36.08 & 114.00 & 20.95  \\
ViT~\cite{dosovitskiy2020vit} & IN-21K~\cite{imagenet_cvpr09} & 35.97 & 113.19 & 20.82 \\
CLIP-I~\cite{Radford2021LearningTV} & 400M~\cite{Radford2021LearningTV} & \textbf{36.96} & \textbf{116.84} & \textbf{21.41} \\
\bottomrule
\end{tabular}
}
\caption{
Image conditioning with image encoded by different pre-trained models. We use the FC method to incorporate the encoded image features.
}
\label{table:img-cond-encoder}
\end{minipage}
\end{table}
In Section~\ref{section:image}, we model the conditional relationship between the detected objects and the input image by an FC layer.
In this subsection, we answer the following questions: (1) What is the better way to model image conditioning? (2) What is the better pre-trained image encoder to encode the input image for image conditioning?
We train the $\mathcal{M}^2$ model with detected objects with different image conditioning methods.
For (1), an alternative is to treat the image features as an additional token, send them into the captioning model together with the set of detected objects, and let the transformer module inside the captioning model learn the conditional relationship.
The image features could be a single token of its global vector representation~\cite{su2019vl,ji2021improving,zhang2021rstnet} (denoted as TF-V) or a sequence of tokens of the grid features (denoted as TF-G).
In Table~\ref{table:img-cond-method}, we see that our proposed simple FC method is the most effective.
Since the computational complexity grows quadratically with respect to the input sequence length for a transformer model, substantially increasing the length of the input sequence (\#Tokens) such as TF-G is undesirable.
By using an FC layer to fuse the features of the detected objects and the input image, we do not increase the length of input tokens.
On the other hand, even though TF-V only increase \#Tokens by one, it underperforms our proposed FC method.

For (2), we also propose to use CLIP-I to encode the input image for image conditioning in Section~\ref{section:image}.
We claim that CLIP-I pre-trained on a similar VL task is capable of preserving as much relevant information from the input image as possible compared to the models pre-trained on image-only datasets such as ImageNet (IN) 1K/21K~\cite{imagenet_cvpr09} or JFT-300M~\cite{sun2017revisiting}.
To verify this claim, we compare image encoded by different image encoders, R-101~\cite{he2016deep}, BiT~\cite{kolesnikov2020big}, ViT~\cite{dosovitskiy2020vit}, and CLIP-I~\cite{Radford2021LearningTV}.
In Table~\ref{table:img-cond-encoder}, we can see that using CLIP-I, which is pre-trained on V+L tasks, as the image encoder performs substantially better than using R-101, BiT, and ViT, which are all pre-trained on image-only datasets.

\textbf{How does image conditioning help.}
In Section~\ref{section:image}, we claim that jointly optimizing the conditional relationship between detected objects and input image helps refine the object features to aid with grounding. %
To verify this, we train the captioning model on the Flickr30k~\cite{plummer2015flickr30k} dataset, which provides the grounding annotations between image and caption pairs.
Following the standard approach as other image captioning papers, we find the most attended object using Integrated Gradient~\cite{sundararajan2017axiomatic} for each word.
Out of 1,014 validation images, our method that refines the object features by image conditioning correctly localizes \textit{\textbf{421 objects}}, while the baseline $\mathcal{M}^2$ only localizes \textit{\textbf{287 objects}}.
More qualitative results can be found in the supplementary.
\section{Conclusion}

In this paper, we address limitations of using pre-trained frozen object detectors as the sole input to auto-regressive models in image captioning. We specifically propose to add an auxiliary branch in the graphical model, leveraging advances in large pre-scaled multi-modal models to retrieve (from the same dataset that the object detector is pre-trained on) \note{contextual attribute and relationship descriptions}. Further, we refine both the detector outputs and retrieved context descriptors in an image-conditioned manner, through a simplified architectural design that avoids significant computational overhead, and show that such conditioning improves quantitative results.
We perform thorough analysis demonstrating that both retrieved text and image conditioning improve results (and jointly even more so), that the multi-modal CLIP model is uniquely suited to our approach, and  grounding improvements.
We also demonstrate significant performance improvements, up to +7.5\% in CIDEr and +1.3\% over already strong state of art.

{\small
\bibliographystyle{ieee_fullname}
\bibliography{sections/egbib}
}

\end{document}